%% file: main.tex
\documentclass[conference, 10 pt, conference]{IEEEtran}  
\usepackage{amsmath,amssymb,amsfonts}
\usepackage{algorithmic}
\usepackage{graphicx}
\usepackage{textcomp}
\usepackage[table,xcdraw]{xcolor}

\usepackage[colorlinks,
            linkcolor=blue,       
            anchorcolor=blue,  
            citecolor=blue,
            urlcolor=blue,        
            ]{hyperref}
\usepackage[table,xcdraw]{xcolor}
\usepackage{rotating}

\usepackage{colortbl}
\usepackage{multirow}
\usepackage{caption}
\usepackage{stfloats}
\usepackage{geometry}
\usepackage{titlesec}

\usepackage[numbers,square,sort]{natbib}

\def\BibTeX{{\rm B\kern-.05em{\sc i\kern-.025em b}\kern-.08em
    T\kern-.1667em\lower.7ex\hbox{E}\kern-.125emX}}

\begin{document}


\title{\vspace{6mm}\LARGE \bf Adaptive Communications in Collaborative Perception with Domain Alignment for Autonomous Driving}

\author{
    \IEEEauthorblockN{Senkang Hu${}^{\star }$, Zhengru Fang${}^{\star}$, Haonan An${}^\ddagger $, Guowen Xu${}^{\star}$, Yuan Zhou$^\ddagger$, Xianhao Chen${}^{\dagger}$, Yuguang Fang${}^{\star}$\\
    $^\star$City University of Hong Kong, Hong Kong, China. Email: \{senkang.forest, zhefang4-c\}@my.cityu.edu.hk,\\ \{guowenxu, my.Fang\}@cityu.edu.hk\\
    {$^\dagger$}The University of Hong Kong, Hong Kong, China. Email: xchen@eee.hku.hk\\
    $^\ddagger$Nanyang Technological University, Singapore. Email: \{an0029an, yzhou027\}@e.ntu.edu.sg
    }
    \vspace{-8mm}
    }

\maketitle

\newcommand\blfootnote[1]{%
\begingroup
\renewcommand\thefootnote{}\footnote{#1}%
\addtocounter{footnote}{-1}%
\endgroup
}


\vspace{-0.4cm}
\begin{abstract}

    \input{sections/abstract.tex}

\end{abstract}


\section{{Introduction}}


Recently, multi-agent collaborative perception \cite{huCollaborativePerceptionConnected2024} has shown a promising solution in autonomous driving to overcome environmental limitations, such as occlusion, extreme weather conditions, and perception range. This kind of perception paradigm allows connected and autonomous vehicles (CAVs) to share their information with others via vehicle-to-everything (V2X) communications, which significantly improve each CAV's perception performance. 

Current approaches aim to strike a balance between performance and bandwidth consumed in communications needed for collaboration. For example, Liu \textit{et al.} \cite{liuWho2comCollaborativePerception2020} employed a multi-step handshake communication process to determine the information of which agents should be shared with.
Liu \textit{et al.} \cite{liuWhen2comMultiAgentPerception2020} developed a communication framework to find the appropriate time to interact with other agents.
Although the aforementioned works on  multi-agent collaborative perception have explored the trade-off between performance and bandwidth \cite{yangWhat2commCommunicationefficientCollaborative2023}, as well as the communication graph construction \cite{liuWho2comCollaborativePerception2020}, these methods all rely on basic proximity-driven design, which fail to consider the impact of dynamic network  capacity on perception performance. 
Unfortunately, wireless channels under vehicular environments are highly dynamic and time-varying, which are affected by the distance between vehicles, the number of vehicles, weather conditions, etc. Without considering the channel variations, the existing works cannot guarantee the transmission rate and delay, which may result in severe performance degradation of perception.
To fill in this gap, we propose a channel-aware strategy to construct the communication graph while minimizing the transmission delay under various channel variations. 

Moreover, to exchange perception data between vehicles while saving bandwidth, prior works employ autoencoders to transmit compressed information, which is then recovered on the receiver side \cite{xuCoBEVTCooperativeBird2022}. However, the existing works along this line  use the basic encoder/decoder techniques, such as  the naive encoder with a single convolutional layer in V2VNet \cite{wangV2VNetVehicletoVehicleCommunication2020} and  a simple $1\times 1$ convolutional auto-encoder in CoBEVT \cite{xuCoBEVTCooperativeBird2022}. These methods cannot meet the transmission latency requirements (e.g., 100ms)  needed for real-time collaborative tasks. To overcome this limitation, we propose an adaptive rate-distortion trade-off strategy with real-time model refinement.

In addition, in many autonomous driving perception tasks, especially in RGB image based tasks, the data heterogeneity across CAVs poses another challenge in collaborative perception. In the real world, different vehicles can encounter different environments during collaborative perception, e.g., one vehicle may be in the dark while another is in the bright spot. Moreover, different types of onboard cameras without unified parameters 
may result in  different variations in the perceived images in brightness, contrast, and color \cite{huFullsceneDomainGeneralization2023}. Therefore, different environments and sensor quality inevitably result in a domain gap between  vehicles, and hence performance degradation in collaborative perception. However, the existing works in collaborative perception design have not considered this effect. To fill in this gap and further improve the performance of collaborative perception, we propose a domain alignment mechanism to reduce the domain gap between different vehicles. The core idea is to transform the images to frequency domain, and then align the amplitude spectrum of the images obtained from different vehicles. 

Following this line of thinking, we propose Adaptive Communications for Collaborative Perception with Domain Alignment (ACC-DA). The core idea is to improve the performance of collaborative perception during both communication and inference phases. Firstly, we take dynamic channel state information (CSI) into consideration to minimize the transmission delay. Secondly, we propose an adaptive rate-distortion trade-off strategy with real-time model refinement. 
These two strategies  can guarantee the effective data sharing and transmission efficiency, which can prevent the performance degradation during communications.
Finally, we propose a domain alignment mechanism to reduce the domain gap between different vehicles, which can further improve the performance during inference. The main contributions of this paper are summarized as follows.
\begin{itemize}
    \item We propose a transmission delay minimization method to construct the communication graph  according to dynamic CSI. 
    \item We develop an adaptive data reconstruction method to not only adjust the rate-distortion trade-off according to CSI but also minimize the temporal redundancy in data during transmission by real-time model refinement to further improve the reconstruction performance.
    \item Finally, we design a domain alignment scheme to reduce the data heterogeneity and domain gap among different CAVs.  This is achieved by transforming the images to frequency domain and align the amplitude spectrum. To our best knowledge, this is the first to consider the domain gap among CAVs in collaborative perception.
\end{itemize}

\section{Related Works}

\subsection{Collaborative Perception}

Even with the significant advances in autonomous driving over the years, 
single-agent perception systems still face significant challenges with occlusions and sensor range limitation. Thus, multi-agent perception has emerged as a solution to effectively tackling these challenges \cite{huWhere2commCommunicationEfficientCollaborative2022}. For example, Wang \textit{et al.} \cite{wangV2VNetVehicletoVehicleCommunication2020} proposed intermediate fusion strategy to enable all agents to only transmit important features derived from the raw point cloud to strike the balance between bandwidth and precision. Li \textit{et al.} \cite{liLearningDistilledCollaboration2021} employed a teacher-student framework to train DiscoGraph via knowledge distillation and proposed  matrix-valued edge weighting to allow an agent to adaptively highlight the informative regions. Xu \textit{et al.} \cite{xuV2XViTVehicletoEverythingCooperative2022} firstly proposed a vision transformer for multi-agent perception and achieved robust performance under location error and communication delay. 
Nevertheless, the aforementioned works have not considered channel variations  and data heterogeneity in vehicular network environments.

\begin{figure*}[t]
    \centering
    \includegraphics[width=1\textwidth]{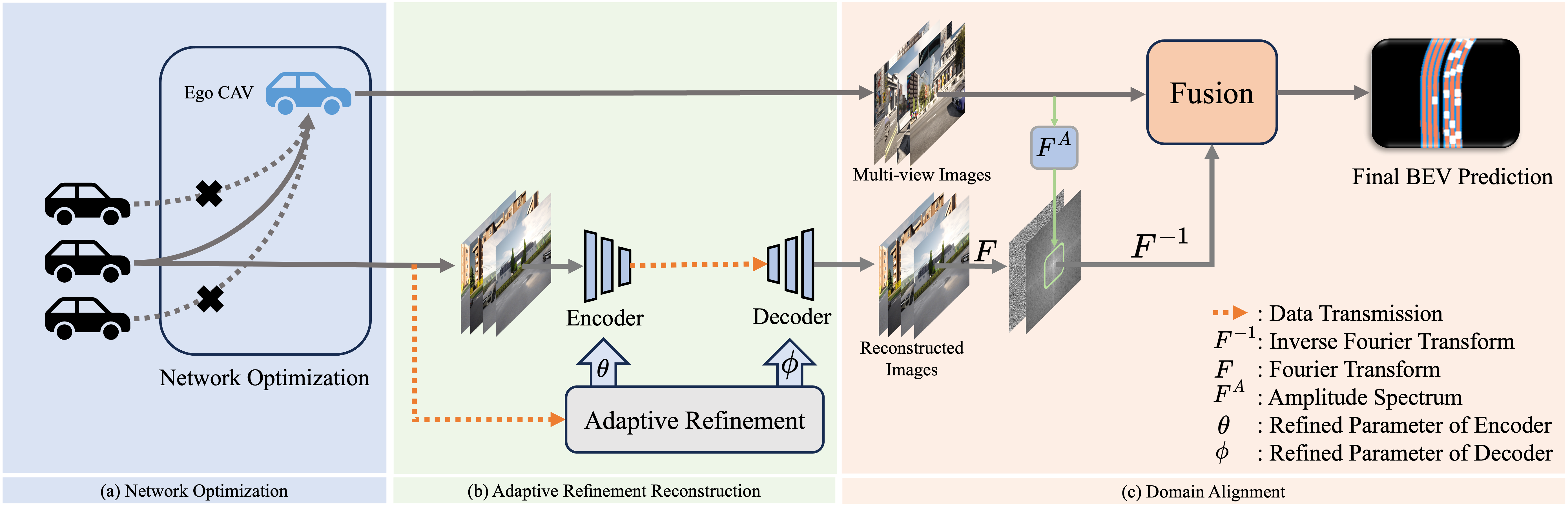}
    \caption{\textbf{Overview architecture} of our proposed ACC-DA framework. First, we minimize the average transmission delay and construct communication graph. Second, CAVs transmit a small portion of raw images to roadside unit to refine the data reconstruction and update the parameters of the encoder and decoder to reduce the temporal redundancy in the data. Meanwhile, CAVs use their encoders to convert images into a bit stream, which is then transmitted to the ego CAV. Third, the ego CAV decodes the received bit stream and aligns the reconstructed images to the domain where its own perceived image in, and then these aligned data are fused together via a fusion net adopted from CoBEVT \cite{xuCoBEVTCooperativeBird2022} to obtain bird's eye view (BEV) prediction. }
    \label{fig:overall-architecture}
    \vspace{-6mm}
\end{figure*}

\subsection{Domain Generalization}
The domain shift problem has seriously impeded large scale deployments of machine learning models \cite{zhouDomainGeneralizationSurvey2022}. To tackle this issue, numerous domain generalization methods have been investigated. 
Domain generalization \cite{zhouDomainGeneralizationSurvey2022} focuses on training a model in multiple source domains so that it can effectively generalize to unfamiliar target domains. 
For example, Li \textit{et al.} \cite{liDomainGeneralizationAdversarial2018} used the autoencoder to minimize the maximum mean discrepancy distance (MMD)
between the distributions of source domains at the feature levels, and they employed adversarial learning \cite{zhouDomainGeneralizationSurvey2022} to ensure that these feature distributions align closely with a predetermined distribution. 
However, our method aims to align the distribution information across CAVs to ego vehicle to reduce the domain shift. Several factors contribute to this choice. First, our method is annotation-free, which is a significant advantage in the context of autonomous driving. Second, our method has the plug-and-use nature, which can be easily integrated into existing training pipelines without any modifications \cite{huFullsceneDomainGeneralization2023}. More details about our domain alignment mechanism will be elaborated in Section \ref{sec:domain-alignment}.

\section{{Proposed Method}}

The goal of our method is to tackle the joint perception issue in autonomous driving.  In this paper, we propose a method called \textbf{A}daptive \textbf{C}ommunications in \textbf{C}ollaborative {P}ercep{t}ion with \textbf{D}omain \textbf{A}lignment ({ACC-DA}), with the overall architecture shown in Fig. \ref{fig:overall-architecture}, consisting of three parts: 1) transmission delay minimization, 2) adaptive data reconstruction, and 3) domain alignment.




\subsection{Transmission Delay Minimization}
\label{sec:channel-aware}

In collaborative perception, transmission delay  serves as a key indicator for CAVs, which is crucial for maintaining  perception accuracy and ensuring safety. It is essential to allow fast data exchange between vehicles, thereby promoting effective data sharing for perception, decision-making, and collaborative actions. To achieve this objective, we propose a transmission delay minimization method to minimize the average transmission delay among CAVs and construct the communication graph.

Let $G \in \mathbb{R}^{N\times N}$ represent the link matrix of a V2V topology communication network of participating  CAVs in collaboration, where $N$ denotes the number of CAVs in collaboration. The link matrix $G$ possesses zero-valued diagonal elements, while all off-diagonal elements are binary. In order to achieve robust communication and remove redundant connections, we should prune the matching matrix 
because communication links with poor channel quality should not be connected. Let $n_j$ denote the ego CAV, and $g_{ij}\in G$ one of the elements of $G$ to represent the transmission link $n_i\rightarrow n_j$. With a constraint of the number of channels $c$, we have:
\begin{equation}
   \sum_i\sum_j g_{ij}\leq c 
   \label{link-constraint}
   \vspace{-2mm}
\end{equation}

Consider that the V2V communication exploits the cellular V2X communication with total bandwidth $W$ equally shared among  $c$ orthogonal sub-channels. The transmission capacity of each sub-channel $C_{ij}$ can be obtained from the Shannon capacity theorem: $C_{ij}=\frac{W}{c}\log_2(1+\frac{P_t h_{ij}}{N_0})$, where
$P_t$ represents the transmit power, $h_{ij}$  the channel gain between the $i$-th transmitter and the $j$-th receiver, and $N_0$ the noise power spectral density.

Moreover, let $T=\{tr_{ij}\}\in \mathbb{R}^{m\times m}$  represent the matrix of transmission rates, where the element $tr_{ij}$ is the amount of data transmitted from vehicle $n_i$ to vehicle $n_j$ per second. Obviously, the transmission rates will not exceed the channel capacity, so we have the constraint of transmission rate $tr_{ij}$:
\begin{equation}
    tr_{ij} \leq C_{ij}
    \label{rate-constraint}
\end{equation}

Let $A_{ij}$ denote the volume of data that vehicle $n_i$ prepares to transmit  to vehicle $n_j$,  
In addition, before transmitting the data, it should be compressed. The transmission delay can then be obtained:
\begin{equation}
    D_{ij}= \gamma_{ij} \cdot A_{ij} / tr_{ij}
    \label{eq: delay}
\end{equation}
where  $\gamma_{ij}\in(0,1]$ is the adaptive compression ratio. 
Considering that the sensing data is obtained from other collaborative CAVs and the data from a closer CAV is more important, which deserves a higher quality of transmission, the adaptive compression ratio $\gamma_{ij}$ can be adjusted according to:
\begin{equation}
    \gamma_{ij}\cdot e^{L_{ij}} \geq \beta
    \label{compression-constraint1}
\end{equation}
where $L_{ij}$ is the Euclidean distance between vehicle $n_i$ and vehicle $n_j$, and $\beta$ is  a fixed constant ($\beta\in(0, 1]$).  Obviously, Eq. (\ref{compression-constraint1}) provides only one way to capture the importance of data exchanged. 

Thus, an optimization problem can be formulated as:
\begin{equation}
    \begin{aligned}
    \underset{\Gamma, G}{\min}   & \sum_{j=1}^N \sum_{i=1}^N g_{i j} D_{i j}  /  \sum_{j=1}^N \sum_{i=1}^N g_{i j}\\
     \text { s.t. }                     &  (\ref{link-constraint}), (\ref{rate-constraint}), (\ref{compression-constraint1})
    \end{aligned}
\end{equation}
In the objective function, we optimize the compression ratio matrix $\Gamma=\{\gamma_{ij}\}_{N\times N}$ and the link matrix $G\in \mathbb{R}^{N\times N}$ to minimize the average transmission delay. Here, $D_{ij}$ is the transmission delay obtained in Eq. (\ref{eq: delay}), $\sum_{j}\sum_{i} g_{i j} D_{i j}$ is the overall time delay in the network, and $\sum_{j} \sum_{i} g_{i j}$ is the total number of transmission links in the network.

To solve this minimization problem, we introduce  Lagrange multipliers $\lambda_{i\in \{1,2,3\}}$ to obtain the Lagrange dual function, hence we can optimize $\Gamma$ and $G$ with gradient descent method as done commonly.


\subsection{Adaptive Model Refinement Reconstruction}

In this section, we propose an adaptive refinement reconstruction method to develop an adaptive rate-distortion (R-D) trade-off strategy with dynamically obtained compression ratio $\gamma_{ij}$ from Sec. \ref{sec:channel-aware}. Additionally, an adaptive model refinement approach has been introduced to further reduce the temporal redundancy in CAV perception data.

Consider an encoder ${y}=f_\theta(x)$ and a decoder $\tilde{{x}} = g_{\phi}(\tilde{y})$, which are convlutional neural networks with parameter $\theta$ and $\phi$, respectively. The model can be trained by minimizing the loss function:
\begin{equation}
    J(\theta,\phi; x)= R(\tilde{y};\theta)+ \beta D(x, \tilde{x};\theta, \phi)
\end{equation}
where $R(\tilde{y};\theta)=E[-\log_2p_{\tilde{y}}(\tilde{y})]$ represents the amount of bits, $D(x, \tilde{x};\theta, \phi)=E[\| x-\tilde{x}\| ^2]$ represents the distortion between the original image $x$ and the reconstructed images $\tilde{{x}}$. 
In order to adaptively adjust the trade-off parameter of R-D $\beta$, we define $\beta_{ij}=\varPhi (\gamma_{ij})$, where the function $\varPhi$ is a non-linear function. For simplicity, $\gamma$ denotes $\gamma_{ij}$ and $\beta$ denotes $\beta_{ij}$.
Then, we can reformulate the traditional fixed rate-distortion problem as a dynamic rate-distortion problem as:
\begin{equation}
    J(\theta,\phi, \gamma; x) = R(\tilde{y};\theta,\gamma)+ \varPhi (\gamma) D(x, \tilde{x};\theta, \phi,\gamma)
\end{equation}
This method allows the dynamic modification of R-D tradeoff based on real-time channel conditions.

Furthermore, to leverage the temporal redundancy of the  successive frames in vehicle-to-vehicle collaborative perception activities, we propose a technique to refine the reconstruction network using a subset of real-time data as the training dataset. We adopt the conditional encoder and decoder presented in \cite{yangVariableRateDeep2020} 
as the backbone (parameters are same) and enhance it with our model refinement strategy. Firstly, the CAV1 sends a portion of raw data to the roadside edge server and then transmit the remaining data with the compression rate $\gamma_{ij}$ to ego CAV. Secondly, the roadside edge server uses the raw data to train the reconstruction network by mean square error minimization. Conceptually, this refinement method enables the model to capitalize on historical data from similar scenarios, enhancing the precision of future image reconstructions.

\subsection{Domain Alignment}
\label{sec:domain-alignment}

\begin{figure}[t]
    \centering
    \includegraphics[width=1\columnwidth]{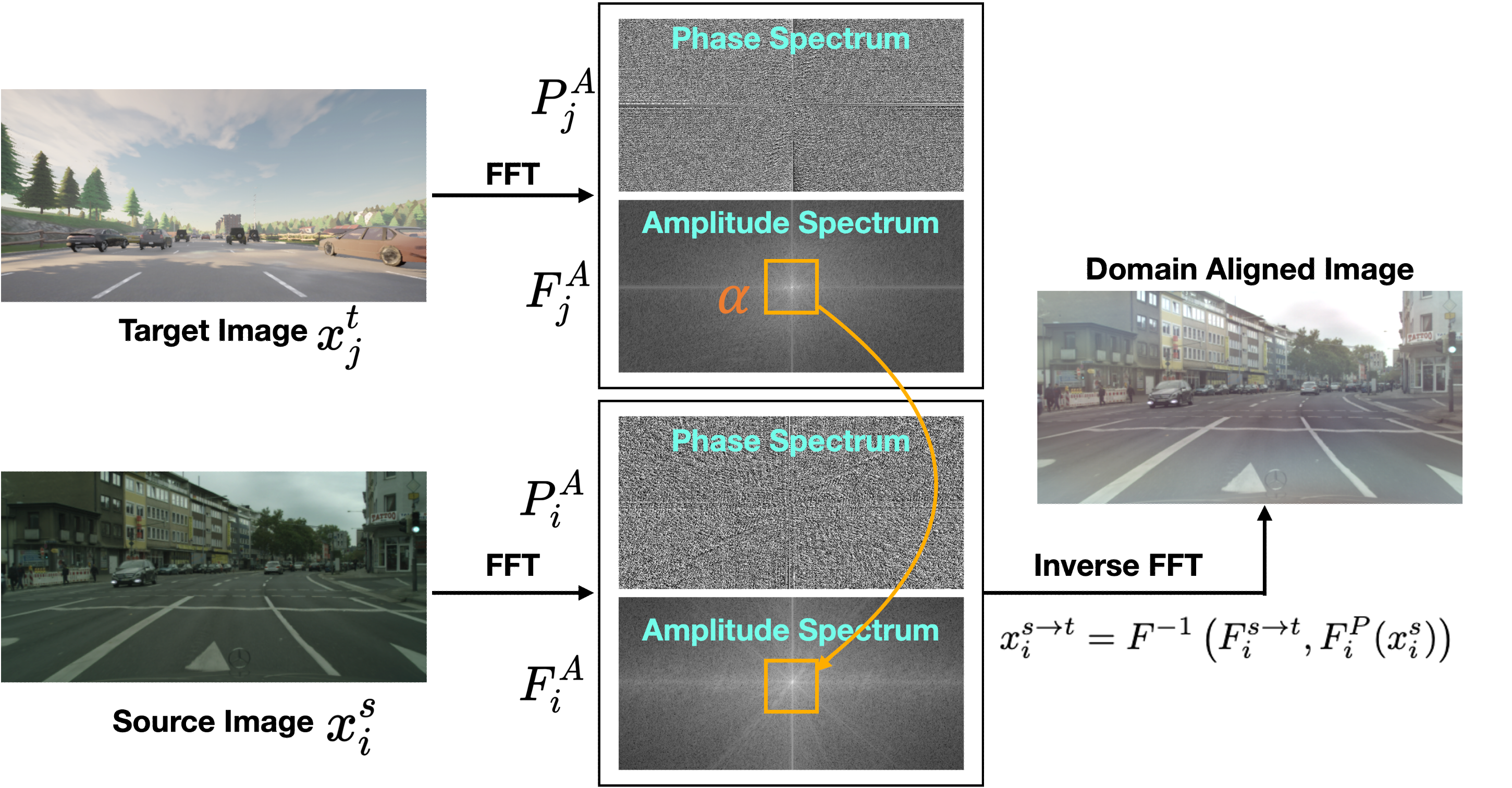}
    \caption{{Domain Alignment (DA) Mechanism}.}
    \label{fig:DA}
    \vspace{-5mm}
\end{figure}

For joint perception in autonomous driving, ego vehicle and other vehicles are situated in different environment, e.g., unbalanced lights: one vehicle in the shaded area and the other in the lighted area. Moreover, different types of car cameras are able to cause chromatic aberration. To tackle this problem, the \textbf{D}omain \textbf{A}lignment \textbf{(DA)} mechanism is proposed. 

Given the dataset $D^t$ of the ego vehicle which is the targeted domain, $D^t=\{x_i^t, y_i^t\}_{i=1}^{N^t}$, where $x_i^t\in \mathbb{R}^{H \times W \times C}$, $C=3$ for RGB image, $C=1$ for grey image, $y_i^t \in \mathbb{R}^{H \times W \times C}$ is the associated label. Similarly $D^s=\{x_i^s, y_i^s\}_{i=1}^{N^s}$ is the source dataset of other collaborative vehicles which we want to align to the target domain.

Specifically, given a sample $x_i$, we can decouple this sample into amplitude $F^A_i\in \mathbb{R}^{H \times W \times C}$ and phase $F^P_i\in \mathbb{R}^{H \times W \times C}$ components by Fourier transform:
\begin{equation}
    F(x_i)(u,v,c)=
    \sum^{H-1}_{j=0}\sum^{W-1}_{k=0}x_i(h,w,c)e^{-2j\pi\left(\frac{h}{H}u+\frac{w}{W}v\right)}
\end{equation}
The amplitude $F^A_i\in \mathbb{R}^{H \times W \times C}$ and phase $F^P_i\in \mathbb{R}^{H \times W \times C}$ represent the low-level distributions (e.g., style) and high-level semantics (e.g., object) of the sample, respectively. Next, in order to reduce, or even eliminate, the domain gap between ego vehicle and other vehicles, we adopt the domain alignment mechanism.

Let a binary mask $M$ be one in the central region, zero in the remaining region: $M(h, w)=\mathbf{1}_{h\times w }$
where $h\in[-\alpha H: \alpha H]$, $ w\in[-\alpha W: \alpha W]$, $\alpha\in (0,1)$. 
Then we generate a new amplitude spectrum distribution by:
\begin{equation}
    \label{efi}
    F_j^{s\rightarrow t}= (I-M)\cdot F_j^A(x^s_j) + M\cdot F_i^A(x^t_i) 
\end{equation}
where $x^s_j$ and $x^t_i$ are randomly sampled from source domain dataset $D^s$ and target domain dataset $D^t$, $I$ is the identity matrix. After obtaining the synthetic amplitude spectrum, we integrate it with the source domain phase spectrum to generate the aligned image by inverse Fourier transform $F^{-1}$. DA mechanism can be formulated as in Eq. (\ref{ex}), and the whole process is shown in Fig. \ref{fig:DA}.
\begin{equation}
    \label{ex}
    x^{s\rightarrow t}_i= F^{-1}\left(F_i^{s\rightarrow t}, F^P_i(x^s_i)\right)
\end{equation}


\begin{figure*}[t]
    \centering
    \includegraphics[width=1\textwidth]{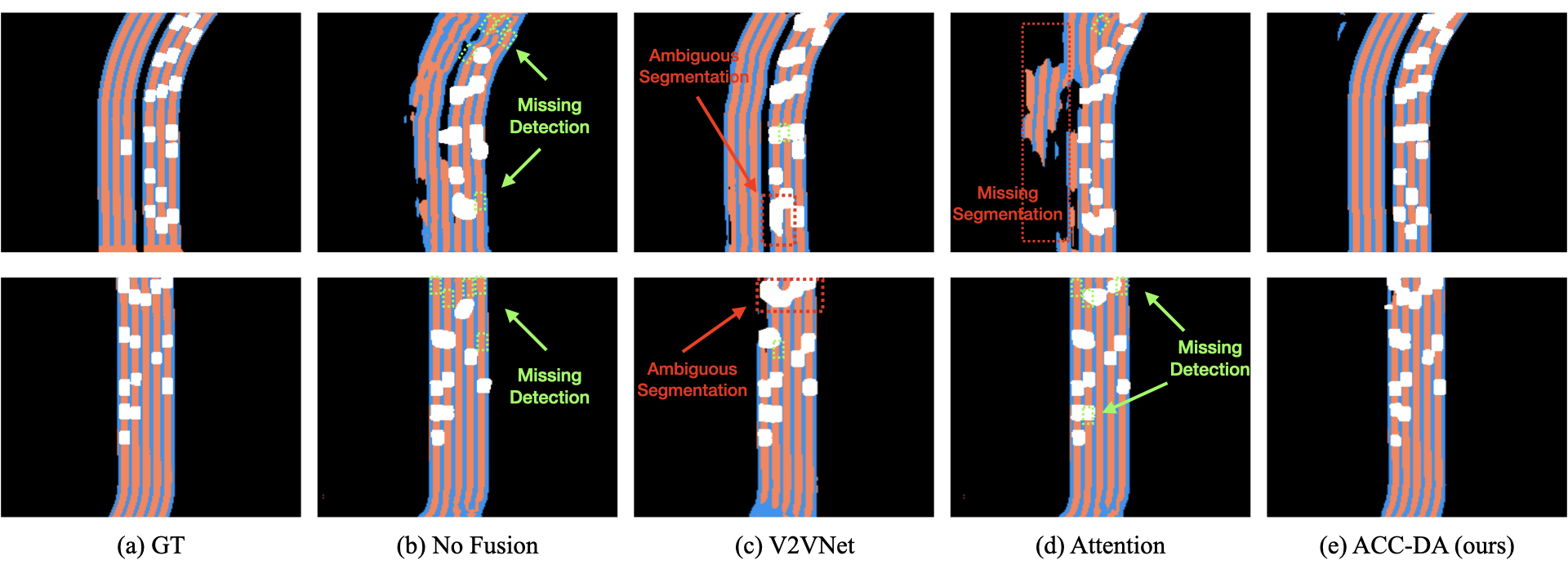}
    \caption{\textbf{Visualization of the BEV segmentation} results from the OPV2V dataset, figure (a) is the Groundtruth, (b) is generated from the No Fusion scheme, (c) is from V2VNet, (d) is from the Attention Fusion. Compared with other methods, our ACC-DA method demonstrates robust performance under different trafﬁc situations, which can achieve more accurate  results.}
    \label{fig:BEV Segmentation}
    \vspace{-6mm}
\end{figure*}

\section{Experiments}

\subsection{Experimental Setting}
\textbf{Dataset and Metrics}. In our experiment, we utilize OPV2V \cite{xuOPV2VOpenBenchmark2022}, a large-scale dataset designed for joint perception with V2V communications. 
It contains 73 different scenarios with an average of 25-seconds duration. To evaluate the performance, we employ the Intersection of Union (IoU) to compare the predicted map against the actual map-view labels, which can be formulated as follows.
\begin{equation}
    IoU = \frac{|BEV_1\bigcap  BEV_0|}{|BEV_1\bigcup BEV_0| }
\end{equation}
where $BEV_0$ represents the ground truth BEV map, and $BEV_1$ represents the predicted BEV map. The IoU is calculated for each class and then averaged across all classes to obtain the final IoU score.

\textbf{Implementation Details.}
Our model is built on PyTorch and trained on two  RTX4090 GPUs utilizing the AdamW optimizer. The initial learning rate is $2\times 10^{-4}$ and decays with an exponential factor of $1\times 10^{-2}$. We employ CoBEVT \cite{xuCoBEVTCooperativeBird2022} as the base to construct our overall architecture.

\subsection{Experimental Results}

\textbf{Perception Performance Evaluation.} To evaluate the BEV segmentation performance of our proposed ACC-DA scheme, we compare it with several existing works, including: 
No Fusion, V2VNet \cite{wangV2VNetVehicletoVehicleCommunication2020}, Attention Fusion \cite{xuOPV2VOpenBenchmark2022}, DiscoNet \cite{liLearningDistilledCollaboration2021}, and CoBEVT \cite{xuCoBEVTCooperativeBird2022}. These schemes assume that the communication channel is ideal as explained before and the domain gap among CAVs is not considered. 
The results are shown in Table \ref{tab:baseline comparison}. Our ACC-DA achieves the best performance in all categories, with an overall IoU of 55.06\%, which is 2.11\% higher than the second-best scheme. In addition, our scheme outperforms the second-best scheme by 4.07\% in terms of vehicle class, which is the most challenging category to differentiate. The results demonstrate that our ACC-DA can effectively improve the perception performance of collaborative perception in autonomous driving.

\begin{table}[t]
    \centering
    \caption{\textbf{Performance Comparison} in Map-view Segmentation on OPV2V Camera-track dataset.}
    \resizebox{1\columnwidth}{!}{
    \begin{tabular}{c|ccccc}
    \hline
    Model        & Road & Lane & Vehicles & Overall  \\ \hline\hline
    No Fusion     & 42.74     & 30.89     &40.73         &   38.12                 \\
    Attention    &   43.30&      31.35          & 45.70               &40.11                          \\
    V2VNet       &     53.00 &         36.11&42.77          & 43.96                         \\
    DiscoNet     &      52.20       &36.19                & 42.97     & 43.48                         \\
    CoBEVT       &      61.78 & 47.65          & 49.43          & 52.95&                 \\
    ACC-DA(ours) &      \textbf{62.60} & \textbf{49.08} & \textbf{53.50} & \textbf{55.06}                          \\ \hline
    \end{tabular}
    }
    \label{tab:baseline comparison}
    \vspace{-4mm}
\end{table}

\textbf{Qualitative Analysis.} To provide a qualitative comparison across different schemes, Fig. \ref{fig:BEV Segmentation} displays the BEV segmentation results  for No Fusion, V2VNet, Attention Fusion, and our ACC-DA across two scenarios, respectively. Evidently, our scheme yields perception results that stand out in terms of both comprehensiveness and accuracy when compared to other schemes. The No Fusion and Attention Fusion schemes exhibit significant omissions both in  vehicles and the road surface. Although V2VNet demonstrates improved outcomes,  it still occasionally misses segments and displays ambiguous boundaries. Most impressively, as observed in Fig. \ref{fig:BEV Segmentation} (e), our scheme excels at almost perfectly segmenting vehicles, road surfaces, and lanes, even for vehicles situated at a considerably further distance (over 100m) from the ego vehicle. The above results show the superiority of our proposed ACC-DA scheme.

\begin{figure}[t]
    \centering
    \includegraphics[width=0.9\columnwidth]{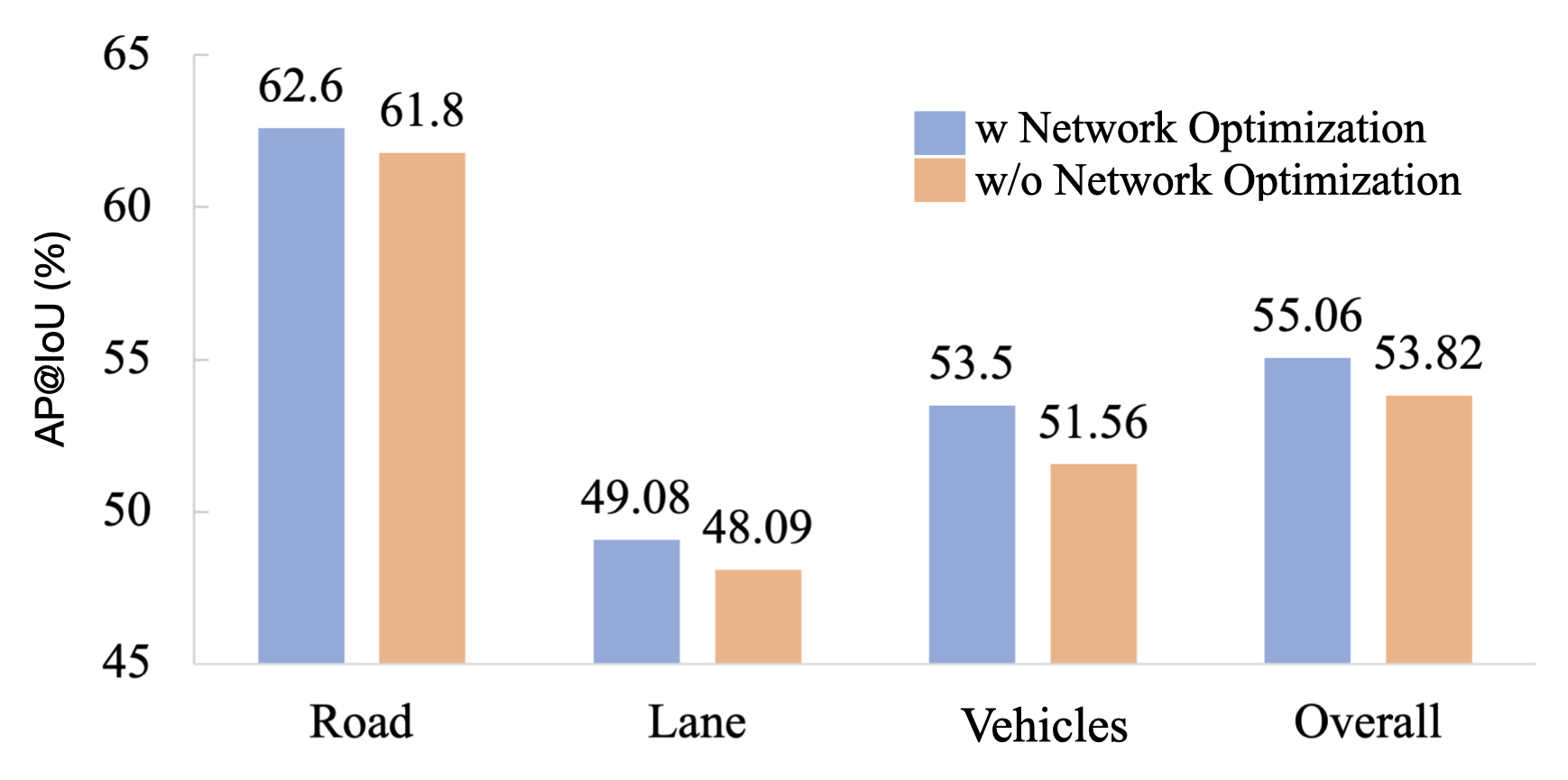}
    \caption{\textbf{Effect of the Network Optimization} "w/" means \textit{with network optimization}, "w/o" means \textit{without network optimization}}
    \label{fig:EffectNetworkOptim}
   \vspace{-5mm}
\end{figure}

\textbf{Effect of Network Optimization.}
In order to evaluate the effect of our transmission delay minimization  discussed in Sec. \ref{sec:channel-aware}, we conduct a comparative experiment depicted in Fig. \ref{fig:EffectNetworkOptim}.  The results reveal that by employing our network optimization, the IoU accuracy of roads, lanes, and vehicles improves by 0.80\%, 1.00\%, and 1.94\%, respectively. Without using our method, the data to the ego vehicle cannot be timely delivered under limited spectrum bandwidth, and the fusion model may incorporate data frames from disparate time instants, resulting in  performance degradation. Notably, the performance in vehicle class is more sensitive to the transmission delay because  vehicles  are more dynamic than roads and lanes. The results demonstrate that our scheme can improve the performance of collaborative perception in autonomous driving by minimizing the transmission delay.

\begin{figure}[t]
    \centering
    \includegraphics[width=0.8\columnwidth]{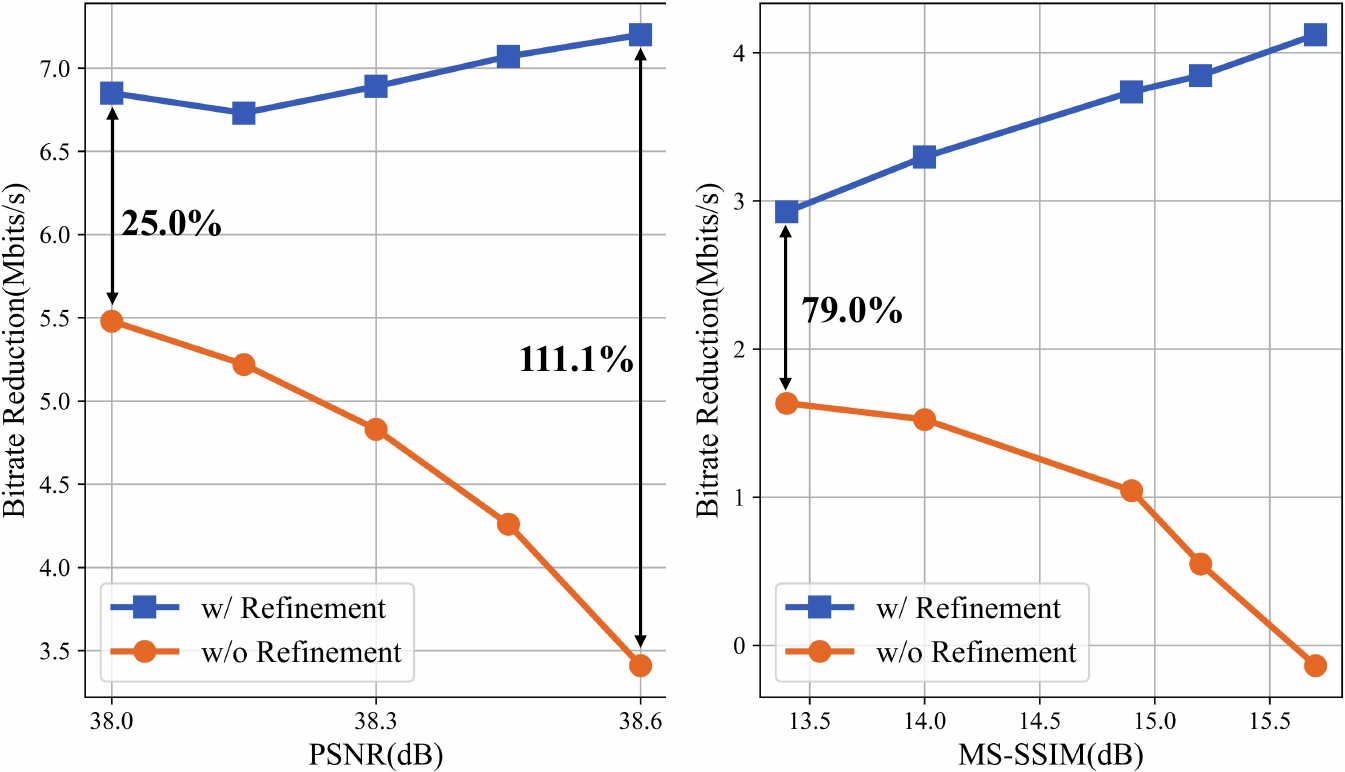}
    \caption{\textbf{Effect of the model refinement.} "w/" means \textit{with refinement}, "w/o" means  \textit{without refinement}}
    \label{fig:bitrate_reduction}
   \vspace{-6mm}
\end{figure}

\textbf{Effect of Adaptive Model Refinement Reconstruction.}
In Fig. \ref{fig:bitrate_reduction}, we conduct an evaluation under two metrics: MS-SSIM
and PSNR
, and we compare the reduction in bitrate achieved by the strategy with and without model refinement. The refinement reconstruction not only achieves a higher reduction in bitrate but also displays an increasing trend as PSNR and MS-SSIM values increase. Conversely, the data reconstruction without model refinement exhibits a decreasing trend. 
Specifically, when compared to the strategy without model refinement, the refined reconstruction leads to a bitrate reduction of 25\% at a PSNR of 38.0 dB, and a reduction of 111.1\% at a PSNR of 38.6 dB.
This significant improvement displays the benefits of the the strategy with model refinement and shows that fune-tuning the parameter distribution of the encoder and decoder through historical data can lead to more efficient image reconstruction.

\begin{figure}[t]
    \centering
    \includegraphics[width=1\columnwidth]{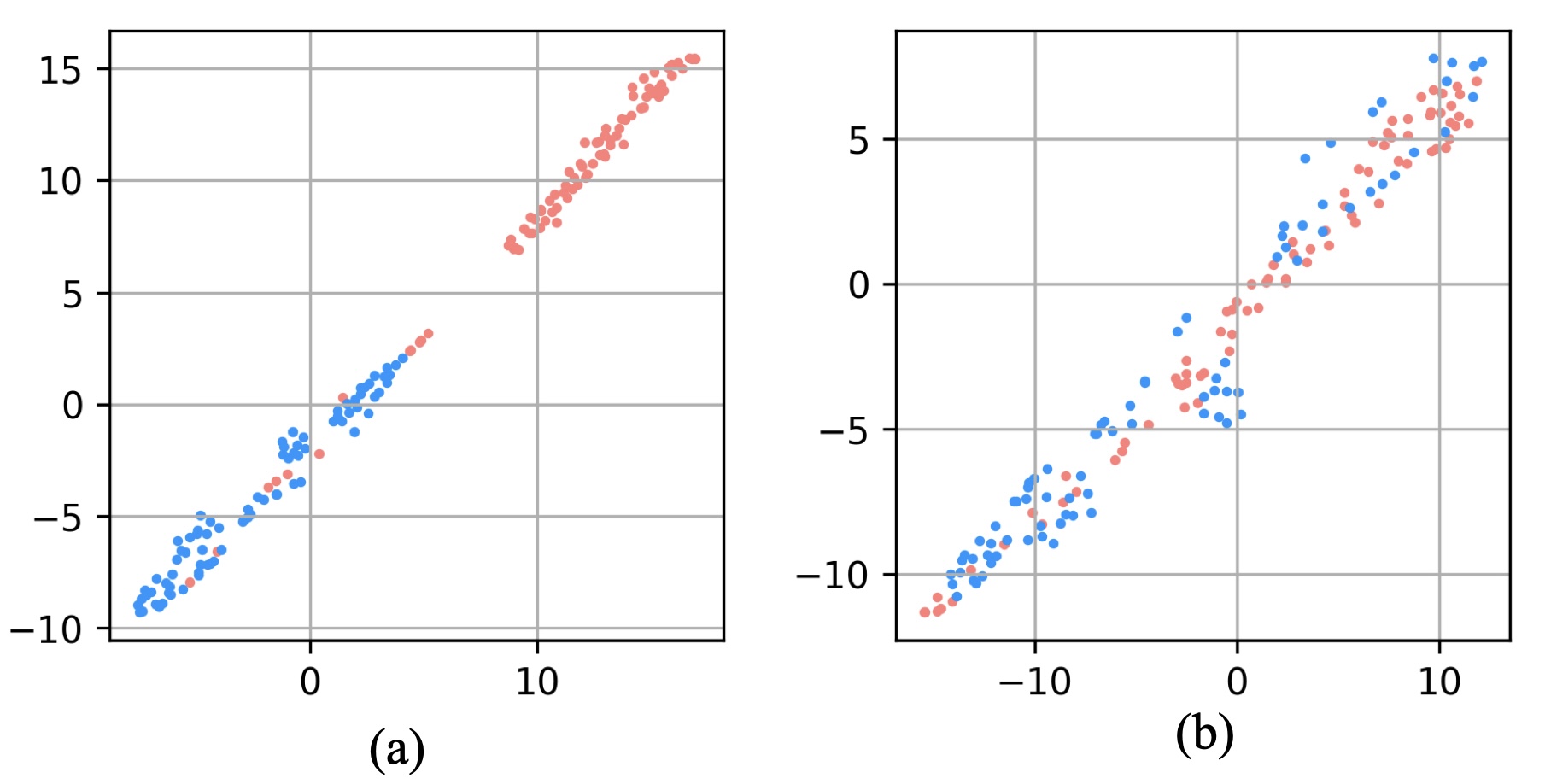}
    \caption{Visualization of t-SNE \cite{van2008visualizing} embedding for the original images (blue) at the ego vehicles and the corresponding images (pink) from other connected vehicles. Left (a) is the original distribution, right (b) is the transformed distribution after domain alignment.}
    \label{fig: t-SNE anslysis}
    \vspace{-6mm}
\end{figure}

\textbf{Effect of Domain Alignment Module}.
In order to study the effect of domain alignment module, we evaluate it in OPV2V dataset with different comparative schemes. 
We present the results in Table \ref{DA-ablation}. As it can be observed, the  domain alignment scheme outperforms all other schemes, especially for the vehicle class, which can lead to an increase of 1 to 3 percentage points in accuracy. Specifically, for the Attention Fusion, DA improves the accuracy by 1.38\%. For DiscoNet and V2VNet, the improvements are 2.10\% and 1.82\%, respectively. The reason is that DA can reduce the distribution heterogeneity among the data in different vehicles. 
To further analyze the impact on the data distribution of DA, we employ t-SNE
\cite{van2008visualizing} to visualize the distribution of images. The result is shown in Fig. \ref{fig: t-SNE anslysis}, where the blue dots represent the distribution of original images at the ego vehicles, and the pink dots represent the corresponding distribution of the images from a different domain in other collaborative connected vehicles. The distribution of the transformed images, as shown in Fig. \ref{fig: t-SNE anslysis} (b), is more concentrated than the original distribution in Fig. \ref{fig: t-SNE anslysis} (a). This indicates that DA can reduce distribution heterogeneity of data in different vehicles, thereby improving joint perception performance.
Overall, our DA can generally improve the performance of vehicle segmentation in camera BEV segmentation, which is significant for joint perception in autonomous driving.

\begin{table}[t]
    \centering
    \caption{\textbf{Ablation Study Results of Domain Alignment}. “w/” means the \textit{with domain alignment}, “w/o” means the \textit{without domain alignment.} }

    \resizebox{1\columnwidth}{!}{
        \begin{tabular}{cc|cccc}
        \hline
        \multicolumn{2}{c|}{AP@IoU}                                  & Road & Lane           & Vehicles       & Overall        \\ \hline\hline
        \multicolumn{1}{c}{\multirow{2}{*}{Attention}}     & w/o DA &43.30 & 31.35          & 45.70               &40.11       \\
        \multicolumn{1}{c}{}                               & w/ DA  &\textbf{43.50} &\textbf{31.53  }        &\textbf{47.78}        & \textbf{40.70}      \\ \hline
        \multicolumn{1}{c}{\multirow{2}{*}{DiscoNet}}     & w/o DA &52.20       &36.19                & 42.97     & 43.48      \\
        \multicolumn{1}{c}{}                               & w/ DA  &\textbf{52.53}  & \textbf{36.57 }        & \textbf{45.07}& \textbf{44.41}              \\ \hline
        \multicolumn{1}{c}{\multirow{2}{*}{V2VNet}}        & w/o DA & 53.00 &         36.11&42.77          & 43.96              \\
        \multicolumn{1}{c}{}                               & w/ DA  &  \textbf{53.03}&          \textbf{36.15}  &\textbf{46.05}    &\textbf{45.08}      \\ \hline
        \multicolumn{1}{c}{\multirow{2}{*}{ACC-DA (ours)}} & w/o DA & 62.85 & 48.97          & 50.42          & 54.08          \\
        \multicolumn{1}{c}{}                               & w/ DA  & 62.60 & \textbf{49.08} & \textbf{53.50} & \textbf{55.06} \\ \hline
        \end{tabular}
    }
    \label{DA-ablation}
    \vspace{-5mm}
\end{table}

\section{Conclusion}

\input{sections/conclusion.tex}

\section{Acknowledgment}
This work is supported in part by the Hong Kong Innovation and Technology Commission (InnoHK Project CIMDA) and the Hong Kong SAR Government under the Global STEM Professorship and Research Talent Hub and the Hong Kong Jockey Club under JC STEM Lab of Smart City.

\footnotesize
{\bibliography{My-Library.bib}

\bibliographystyle{IEEEtran}}

\end{document}

%% file: sections/abstract.tex
Collaborative perception among multiple connected and autonomous vehicles (CAVs)  can greatly enhance perceptive capabilities by allowing vehicles to exchange supplementary information. Despite
significant advances, many design challenges still remain
due to channel variations and data heterogeneity among
collaborative vehicles. To address these issues, we propose
ACC-DA, a channel-aware collaborative perception framework to
dynamically adjust the communication graph to minimize the
average transmission delay while mitigating the impacts caused by data heterogeneity. 
More specifically, we first 
construct the communication graph to minimize the transmission delay
according to different channel information state. We then propose
an adaptive data reconstruction mechanism to dynamically adjust the rate-distortion trade-off to enhance perception efficiency 
while reducing the temporal redundancy during data transmissions.
Finally,  we conceive a domain alignment scheme  to align
the data distribution from different vehicles to 
mitigate the domain gap between different vehicles and improve
the performance of the target task. Comprehensive experiments
demonstrate the effectiveness of our method in comparison to the existing state-of-the-art works.

%% file: sections/conclusion.tex
In this paper, we have developed {ACC-DA}, a novel multi-agent perception framework, which includes three modules: i) transmission delay minimization module to dynamically adjust the communication graph and minimize the average transmission delay among CAVs, ii) adaptive refinement reconstruction module to  adjust the R-D trade-off and reduce the temporal redundancy in data to improve the transmission efficiency, and iii) domain alignment module to handle the data heterogeneity between different collaborative vehicles to further enhance the perception performance and reliability. Comprehensive experiments verify the superiority of our framework compared with the existing state-of-the-art methods.
